\documentclass{article}

\usepackage{microtype}
\usepackage{graphicx}
\usepackage{subfigure}
\usepackage{booktabs} 

\usepackage{hyperref}
\usepackage{caption}

\usepackage[accepted]{icml2025}

\usepackage{amsmath}
\usepackage{amssymb}
\usepackage{mathtools}
\usepackage{amsthm}
\usepackage{subcaption}
\usepackage[capitalize,noabbrev]{cleveref}
\usepackage{textcomp}

\theoremstyle{plain}

\theoremstyle{definition}

\theoremstyle{remark}

\usepackage[textsize=tiny]{todonotes}

\icmltitlerunning{Submission and Formatting Instructions for ICML 2025}

\begin{document}

\twocolumn[
\icmltitle{
Towards Interpretable Time Series Foundation Models
}




\begin{icmlauthorlist}
\icmlauthor{Matthieu Boileau}{1}
\icmlauthor{Philippe Helluy}{1}
\icmlauthor{J\'er\'emy Pawlus}{2}
\icmlauthor{Svitlana Vyetrenko}{3,1}
\end{icmlauthorlist}

\icmlaffiliation{1}{Univeristy of Strasbourg, France}
\icmlaffiliation{2}{AxesSim, Strasbourg, France}
\icmlaffiliation{3}{Outsampler, Strasbourg, France}

\icmlcorrespondingauthor{Svitlana Vyetrenko}{svitlana@outsampler.com}

\icmlkeywords{Machine Learning, ICML}

\vskip 0.3in
]



\printAffiliationsAndNotice{}  

\begin{abstract}
In this paper, we investigate the distillation of time series reasoning capabilities into small, instruction-tuned language models as a step toward building interpretable time series foundation models. Leveraging a synthetic dataset of mean-reverting time series with systematically varied trends and noise levels, we generate natural language annotations using a large multimodal model and use these to supervise the fine-tuning of compact \texttt{Qwen} models. We introduce evaluation metrics that assess the quality of the distilled reasoning—focusing on trend direction, noise intensity, and extremum localization—and show that the post-trained models acquire meaningful interpretive capabilities. Our results highlight the feasibility of compressing time series understanding into lightweight, language-capable models suitable for on-device or privacy-sensitive deployment. This work contributes a concrete foundation toward developing small, interpretable models that explain temporal patterns in natural language.
\end{abstract}

\section{Introduction}
\subsection{Problem statement}
\label{problem}
Recent research has shown that small models can achieve performance comparable to much larger models when trained on carefully curated datasets. For instance, \cite{textbooksneed} demonstrate that high-quality, instruction-tuned data can dramatically improve model capabilities without scaling up parameters. Similarly,  \cite{tinystories} highlights that small language models trained on compact, well-structured data can generate coherent and creative narratives. This principle is further reinforced by advancements like \texttt{\small SKY T1} \cite{sky_t1_2025} and \texttt{\small DeepSeek R1} \cite{deepseekr1}, where focused data selection and training techniques enable relatively lightweight models to perform tasks traditionally reserved for larger architectures. Together, these studies suggest that data quality, not just model size, is a critical factor in achieving strong language model performance.

Time series data is ubiquitous across healthcare, finance, and industrial systems, where real-time, on-device analysis is essential for operational efficiency and timely decision-making. Models in these settings must address a range of tasks, including forecasting, anomaly detection, trend analysis, and event classification. Beyond performance and compactness, it is increasingly important for such models to be interpretable—able to explain temporal patterns and anomalies in natural language that aligns with human reasoning. Despite its significance, this direction remains largely unexplored. In this work, we take a step toward this goal by introducing a method for distilling time series reasoning into small, interpretable models capable of generating structured natural language explanations.

\subsection{Related work}
\label{related}
Recent work has investigated the application of general-purpose language models to time series tasks. Gruver and Wilson \cite{gruver2024largelanguagemodelszeroshot} first demonstrated that pretrained language models can achieve competitive zero-shot performance on standard forecasting benchmarks. Building on this, Jin et al. \cite{jin2024positionlargelanguagemodels} proposed a broader framework in which a variety of time series tasks are reformulated as token prediction problems within a language modeling paradigm. In parallel, specialized foundational models pretrained exclusively on time series data have been introduced. Chronos \cite{chronos2024}, Llaglama \cite{lagllama}, and TimesFM \cite{timesfm} focus on large-scale pretraining for forecasting, achieving strong zero-shot and few-shot performance across diverse temporal datasets. Moirai \cite{moirai} extends these methods to multivariate forecasting, capturing complex dependencies among multiple correlated series. Multitask models such as UniTS \cite{units} and Moment \cite{moment} further generalize this approach by jointly training across forecasting, imputation, and classification tasks, illustrating the advantages of shared temporal representations.

Systematic evaluations based on the series feature taxonomy proposed in \cite{fons2024evaluatinglargelanguagemodels} benchmark general-purpose language models on a range of time series understanding tasks, highlighting both their strengths and limitations relative to specialized models. TimeSeriesExam \cite{cai2024timeseriesexamtimeseriesunderstanding} introduces a complementary benchmark focused on evaluating language models' ability to perform detailed reasoning over temporal patterns. In parallel, work in \cite{googlechartbaselines} extends language model-based methods to the interpretation of visual temporal data, including line and bar charts. Together, these efforts reflect a broader shift toward treating time series and structured data as instances of language modeling.


\subsection{Our contributions}
\label{contributions}
\begin{enumerate}
    \item We propose a framework for {\bf interpretable time series foundational model} construction via knowledge distillation, enabling small language models to reason about time series data in natural language.
    \item We introduce a {\bf practical and reproducible distillation pipeline} that transfers simple time series reasoning skills—such as trend identification, noise assessment, and extrema localization —- from large multimodal models to compact ones.
    \item We contribute a {\bf baseline dataset and evaluation framework} for assessing time series reasoning in small models, providing a foundation for future work on interpretable time series foundation models.
\end{enumerate}

\section{Distillation of time series reasoning}

\subsection{Background} Distillation is the process of transferring knowledge from a large, often complex model (the teacher) to a smaller, more efficient model (the student) that can understand and reason about a narrow topic \cite{xu2024surveyknowledgedistillationlarge, hinton2015distillingknowledgeneuralnetwork}. This technique aims to retain the performance of the larger model while reducing computational requirements, making deployment more practical in real-time or resource-constrained settings. In addition, practical benefits can include the ability to fine-tune smaller models with sensitive data. 

Recent advances in distilling reasoning—particularly mathematical inference—into small models have demonstrated that compact architectures can tackle complex tasks with surprising accuracy\cite{sky_t1_2025, deepseekr1}. Distillation of time series reasoning has not been explored yet in the literature. In this paper, we explore this capability and demonstrate that small models can be effectively trained via distillation.

\subsection{Synthetic dataset generation and annotation}
We create an annotated dataset of time series given by Orstein-Uhlenbeck process \cite{byrd2019explainingagentbasedfinancialmarket}:
\begin{align}
 r_{t}=r_{t-1}+\kappa(\overline{r}-r_{t-1})+u_{t},\quad r_0 = 0,
\label{equation}
\end{align}
where $\bar{r}$ is a mean value of the process, $\kappa$ is a mean-reversion parameter and $u_t \sim \mathcal{N} (0, \sigma^2)$ is random noise added to the time series at each time step $t$.

\begin{figure}
\begin{subfigure}
{\includegraphics[width=0.45\textwidth]{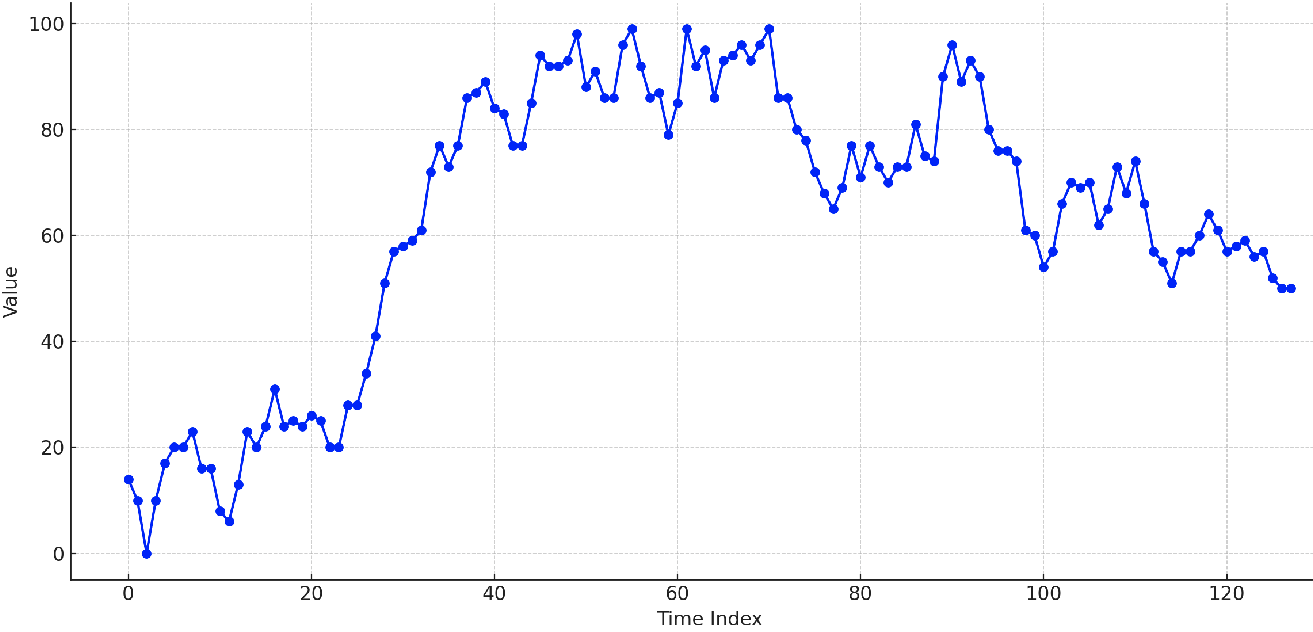}}
\caption{Sample time series image input to {\small \texttt{pixtral-large}}.}
\label{ts_image}
\end{subfigure}
\begin{subfigure}
{\colorbox{pink}{\parbox{8cm}{PROMPT: {\small Describe the time series in three sentences. First sentence: describe trend (increasing/decreasing/flat). Second sentence: noise intensity (low/medium/high). Third sentence: approximate localisation of global maximum (beginning/middle/end) and global minimum (beginning/middle/end).
Put the description in a JSON format with the following pattern
\{ "trend": \textless{} sentence1 \textgreater{},
  "noise": \textless{} sentence2 \textgreater{},
  "extrema": \textless{} sentence3 \textgreater{} \}
}}}}
\caption{Prompt used to annotate time series by {\small \texttt{pixtral-large}}. It is worth noting that such prompting is requiring the model to simply localize the extrema, without explicitly specifying them.}
\label{ts_prompt}
\end{subfigure}
\begin{subfigure}
{\colorbox{lightgray}{\parbox{8cm}{ANNOTATION: {\small The time series exhibits an increasing trend initially, followed by fluctuations and a general decreasing trend towards the end. The noise intensity in this time series is high, with significant fluctuations throughout. The global maximum is approximately located in the middle of the time series, while the global minimum is towards the beginning.}}}}
\caption{Sample time series annotation generated by {\small \texttt{pixtral-large}}.}
\label{ts_output}
\end{subfigure}
\end{figure}

Orstein-Uhlenbeck process is frequently used to model financial markets \cite{wah_wellman, bamford2024multimodalfinancialtimeseriesretrieval} as well as biological processes \cite{biological}.
As argued in 
 \cite{zhou2025llmsunderstandtimeseries}, large language models generally perform  better in time series reasoning when processing time series data as images rather than text tokens; therefore, we use both numerical and visual time series presentations in order to generate high-quality time series annotations by a large model. 
We  generate time series captions by prompting a large model {\small \texttt{pixtral-large}} by Mistral AI to generate three sentences that describe the time series trend (further referred to as "trend"), possible presence and intensity of noise (further referred to as "noise") and description of local and global extrema (further referred to as "extrema"). See Figures~\ref{ts_image}, ~\ref{ts_prompt} and ~\ref{ts_output} for the sample time series image input, prompt and the generated output. Using the above procedure, and varying randomly the parameters $(\kappa, \overline{r}, \sigma)$ we generate 200 time series-image-annotation samples. Then we will allocate 180 samples for training, and 20 samples for testing.

\subsection{Evaluation metrics}
\label{evaluation_metrics}

In order to evaluate the quality of the dataset and, subsequently, the quality of the post-trained small models, we propose the following evaluation metrics:
\begin{enumerate}
    \item {\bf{Standard cosine similarity metric}} between the entire annotated paragraphs (that is, the three sentences) computed with the sentence-transformers model {\small \texttt{all-MiniLM-L6-v2}}. Cosine similarity score is known to measure overall relatedness and syntactic similarity in the annotation style, however, might not be able to capture a strong contradiction in the logical meaning (e.g., "Time series is increasing" and "Time series is decreasing").
    \item {\bf{NLI scores}} on a per-sentence level computed with {\small \texttt{roberta-large-mnli}} model. Unlike cosine similarity score above, NLI scores help highlight if there is a strong contradiction between the generated annotations and the ground truth \cite{williams2018broadcoveragechallengecorpussentence}.
    For each of the three diagnostic sentences (“trend”, “noise”, “extrema”), we attribute 0 to contradiction, 0.5 to neutral and 1 to entailment and then compute the average score across the samples. 
    \item {\bf{Feature-based similarity scores}} We assess the quality of annotations by comparing them to fact-based reference sentences computed based on explicit derivation of time series features using the following procedure. First, the trend is determined based on the overall increasing or decreasing behavior of a smoothed version of the time series. Second, the noise level is estimated from the variance of the data. Third, the positions of global extrema are computed precisely in order to describe their approximate locations. These three analyses are used to generate fact-checked sentences such as: "The time series shows an overall increasing/flat/decreasing trend"; "The noise intensity is low/medium/high"; "The minimum/maximum occurs around the beginning/middle/end of the time series".
\end{enumerate}

\subsection{Fact-checking}
\label{section_fact_check}To improve the confidence in our dataset, we perform a NLI comparison between {\small \texttt{pixtral-large}} and our feature-based annotations.
For the "trend" and "noise" annotations, we observe agreement rates of 93\% and 95\%, respectively, between {\small \texttt{pixtral-large}} and the fact-based sentences. The few detected discrepancies generally reflect acceptable ambiguity. For example, a time series with large fluctuations might reasonably be described as either "increasing" or exhibiting "no clear trend." We choose to retain such cases in the dataset. No unambiguous contradictions (such as simultaneous claims of "increasing" and "decreasing" trends) were observed, which aligns with the documented performance of {\small \texttt{pixtral-large}} on the ChartQA benchmark~\cite{pixtral-large}.

For the "extrema" annotation, the NLI comparison identifies 5\% of cases as contradictions. Some of these reflect genuine inconsistencies. In such cases, we replace the annotation with the corresponding fact-based sentence to ensure the overall quality of the dataset.


\subsection{Small model training}

We use the curated dataset to fine-tune \texttt{\small Qwen2.5-1.5B-Instruct} and \texttt{\small Qwen2.5-0.5B-Instruct} models \cite{qwen2025qwen25technicalreport}. Such models are equipped with basic natural language capabilities, however, due to their size they are expected to perform worse on time series reasoning without post-training (as our subsequent testing then confirms - see Table~\ref{table_before_training}). At this stage, we fine-tune the small \texttt{Qwen} models using only the numerical time series data paired with their annotated text, and without including time series images. 

It should be noted that tokenization of numerical values poses challenges for large language models (LLMs), as most tokenizers are optimized for natural language and not for numeric precision \cite{spathis2023stephardestpitfallsrepresenting}. Decimal numbers, scientific notation, and long integers are often split into multiple tokens inconsistently, which can hinder model performance on tasks involving quantitative reasoning. To ensure consistent tokenization, we rescale and round all floating numbers obtained by the time series generation process to integers with left padding from \text{00} to {99}, and ensure that each digit is represented as a single token \cite{yuan2023largelanguagemodelsperform}.

\subsection{Experimental results and evaluation}
To evaluate the quality of time series reasoning that we transfer to small models \texttt{\small Qwen2.5-1.5B-Instruct} and \texttt{\small Qwen2.5-0.5B-Instruct} through the distilled samples, we compare the annotations they generate before and after post-training with the initial annotations generated by {\small \texttt{pixtral-large}} on the test dataset using the evaluation metrics described in Section~\ref{evaluation_metrics}. 

As evident from Table~\ref{table_before_training}, we observe that initially the small \texttt{\small Qwen} models do not display much time series reasoning. Notably, small models understand trends better than other features which agrees with test results for large models in \cite{fons2024evaluating} and \cite{cai2024timeseriesexamtimeseriesunderstanding}. The experimental results were significantly improved after post-training as one can observe in Table~\ref{table_after_training}. Therefore, our experiments suggest that even very small models are capable of learning basic time series reasoning in natural language.

\begin{center}
\begin{table}
\begin{tabular}{ |c|c|c| } 
 \hline
 & {\small\texttt{Qwen2.5-1.5B}} & {\small\texttt{Qwen2.5-0.5B}}  \\ 
 \hline
 Cosine & 0.5 & 0.62 \\ 
 \hline
 NLI "trend" & 0.7 & 0.6 \\ 
 \hline
 NLI "noise" & 0.575 & 0.275 \\ 
 \hline
 NLI "extrema" & 0.35 & 0.475 \\ 
 \hline
 Feature "trend" & 0.8 & 0.6 \\ 
 \hline
 Feature "noise" & 0.475 & 0.45 \\ 
 \hline
 Feature "extrema" & 0.475 & 0.45 \\ 
 \hline
\end{tabular}
\caption{Evaluation results before post-training. Note that prior to training, the small models are unable to produce valid JSON outputs in the expected three-sentence format. Therefore, for the pre-training evaluation, we assess their performance by asking separate questions about trend, noise, and extrema, without requiring JSON formatting.}
\label{table_before_training}
\end{table}
\begin{table}
\begin{tabular}{ |c|c|c| } 
 \hline
 & {\small\texttt{Qwen2.5-1.5B}} & {\small\texttt{Qwen2.5-0.5B}} \\ 
 \hline
 Cosine & 0.98 & 0.98 \\ 
 \hline
 NLI "trend" & 0.875 & 0.875 \\ 
 \hline
 NLI "noise" & 0.75 & 0.675 \\ 
 \hline
 NLI "extrema" & 0.875 & 0.775 \\ 
 \hline
 Feature "trend" & 1.0 & 1.0 \\ 
 \hline
 Feature "noise" & 0.8 & 0.8 \\ 
 \hline
 Feature "extrema" & 0.95 & 0.95 \\ 
 \hline
\end{tabular}
\caption{Evaluation results after post-training}
\label{table_after_training}
\end{table}
\end{center}

\section{Conclusion}

Our findings demonstrate that distillation of time series reasoning is feasible even for very small models, enabling effective downstream use in resource-constrained environments such as wearable devices. Such post-trained models can be further fine-tuned with application-specific data, and they exhibit promising utility in real-world scenarios. 

Our future work will focus on enhancing interpretable foundation model for time series capabilities through visual and numerical representation learning, exploring alternative numerical tokenization strategies for effective floating-point representation, and also improving precision in arithmetic reasoning and retrieval. Additionally, small post-trained models equipped with time-series specific reasoning capability offer a useful testbed for observing and understanding the out-of-training-distribution generalization, which we plan to explore in greater depth in our future work.

\section*{Acknowledgements} This work has been supported by the Institut Thematique Interdisciplinaire IRMIA++ at the University of Strasbourg (\url{https://irmiapp.unistra.fr/}) and the Gutenberg Circle.

\section*{Impact Statement}

This paper presents work whose goal is to advance the field of 
Machine Learning. There are many potential societal consequences 
of our work, none which we feel must be specifically highlighted here.


\bibliography{custom}

\begin{thebibliography}{28}
\providecommand{\natexlab}[1]{#1}
\providecommand{\url}[1]{\texttt{#1}}
\expandafter\ifx\csname urlstyle\endcsname\relax
  \providecommand{\doi}[1]{doi: #1}\else
  \providecommand{\doi}{doi: \begingroup \urlstyle{rm}\Url}\fi

\bibitem[pix()]{pixtral-large}
Available from \url{https://mistral.ai/news/pixtral-large}.

\bibitem[Ansari et~al.()Ansari, Stella, Turkmen, Zhang, Mercado, Shen, Shchur, Rangapuram, Arango, Kapoor, Zschiegner, Maddix, Wang, Mahoney, Torkkola, Wilson, Bohlke-Schneider, and Wang]{chronos2024}
Ansari, A.~F., Stella, L., Turkmen, C., Zhang, X., Mercado, P., Shen, H., Shchur, O., Rangapuram, S.~S., Arango, S.~P., Kapoor, S., Zschiegner, J., Maddix, D.~C., Wang, H., Mahoney, M.~W., Torkkola, K., Wilson, A.~G., Bohlke-Schneider, M., and Wang, Y.
\newblock Chronos: Learning the language of time series.

\bibitem[Bamford et~al.(2024)Bamford, Coletta, Fons, Gopalakrishnan, Vyetrenko, Balch, and Veloso]{bamford2024multimodalfinancialtimeseriesretrieval}
Bamford, T., Coletta, A., Fons, E., Gopalakrishnan, S., Vyetrenko, S., Balch, T., and Veloso, M.
\newblock Multi-modal financial time-series retrieval through latent space projections, 2024.
\newblock URL \url{https://arxiv.org/abs/2309.16741}.

\bibitem[Byrd(2019)]{byrd2019explainingagentbasedfinancialmarket}
Byrd, D.
\newblock Explaining agent-based financial market simulation, 2019.
\newblock URL \url{https://arxiv.org/abs/1909.11650}.

\bibitem[Cai et~al.(2024)Cai, Choudhry, Goswami, and Dubrawski]{cai2024timeseriesexamtimeseriesunderstanding}
Cai, Y., Choudhry, A., Goswami, M., and Dubrawski, A.
\newblock Timeseriesexam: A time series understanding exam, 2024.
\newblock URL \url{https://arxiv.org/abs/2410.14752}.

\bibitem[Das et~al.(2024)Das, Kong, Sen, and Zhou]{timesfm}
Das, A., Kong, W., Sen, R., and Zhou, Y.
\newblock A decoder-only foundation model for time-series forecasting, 2024.
\newblock URL \url{https://arxiv.org/abs/2310.10688}.

\bibitem[Daswani et~al.(2024)Daswani, Bellaiche, Wilson, Ivanov, Papkov, Schnider, Tang, Lamerigts, Botea, Sanchez, Patel, Prabhakara, Shetty, and Telang]{googlechartbaselines}
Daswani, M., Bellaiche, M. M.~J., Wilson, M., Ivanov, D., Papkov, M., Schnider, E., Tang, J., Lamerigts, K., Botea, G., Sanchez, M.~A., Patel, Y., Prabhakara, S., Shetty, S., and Telang, U.
\newblock Plots unlock time-series understanding in multimodal models, 2024.
\newblock URL \url{https://arxiv.org/abs/2410.02637}.

\bibitem[DeepSeek-AI et~al.(2025)DeepSeek-AI, Guo, Yang, Zhang, Song, Zhang, Xu, Zhu, Ma, Wang, Bi, Zhang, Yu, Wu, Wu, Gou, Shao, Li, Gao, Liu, Xue, Wang, Wu, Feng, Lu, Zhao, Deng, Zhang, Ruan, Dai, Chen, Ji, Li, Lin, Dai, Luo, Hao, Chen, Li, Zhang, Bao, Xu, Wang, Ding, Xin, Gao, Qu, Li, Guo, Li, Wang, Chen, Yuan, Qiu, Li, Cai, Ni, Liang, Chen, Dong, Hu, Gao, Guan, Huang, Yu, Wang, Zhang, Zhao, Wang, Zhang, Xu, Xia, Zhang, Zhang, Tang, Li, Wang, Li, Tian, Huang, Zhang, Wang, Chen, Du, Ge, Zhang, Pan, Wang, Chen, Jin, Chen, Lu, Zhou, Chen, Ye, Wang, Yu, Zhou, Pan, Li, Zhou, Wu, Ye, Yun, Pei, Sun, Wang, Zeng, Zhao, Liu, Liang, Gao, Yu, Zhang, Xiao, An, Liu, Wang, Chen, Nie, Cheng, Liu, Xie, Liu, Yang, Li, Su, Lin, Li, Jin, Shen, Chen, Sun, Wang, Song, Zhou, Wang, Shan, Li, Wang, Wei, Zhang, Xu, Li, Zhao, Sun, Wang, Yu, Zhang, Shi, Xiong, He, Piao, Wang, Tan, Ma, Liu, Guo, Ou, Wang, Gong, Zou, He, Xiong, Luo, You, Liu, Zhou, Zhu, Xu, Huang, Li, Zheng, Zhu, Ma, Tang, Zha, Yan, Ren, Ren, Sha, Fu, Xu, Xie, Zhang,
  Hao, Ma, Yan, Wu, Gu, Zhu, Liu, Li, Xie, Song, Pan, Huang, Xu, Zhang, and Zhang]{deepseekr1}
DeepSeek-AI, Guo, D., Yang, D., Zhang, H., Song, J., Zhang, R., Xu, R., Zhu, Q., Ma, S., Wang, P., Bi, X., Zhang, X., Yu, X., Wu, Y., Wu, Z.~F., Gou, Z., Shao, Z., Li, Z., Gao, Z., Liu, A., Xue, B., Wang, B., Wu, B., Feng, B., Lu, C., Zhao, C., Deng, C., Zhang, C., Ruan, C., Dai, D., Chen, D., Ji, D., Li, E., Lin, F., Dai, F., Luo, F., Hao, G., Chen, G., Li, G., Zhang, H., Bao, H., Xu, H., Wang, H., Ding, H., Xin, H., Gao, H., Qu, H., Li, H., Guo, J., Li, J., Wang, J., Chen, J., Yuan, J., Qiu, J., Li, J., Cai, J.~L., Ni, J., Liang, J., Chen, J., Dong, K., Hu, K., Gao, K., Guan, K., Huang, K., Yu, K., Wang, L., Zhang, L., Zhao, L., Wang, L., Zhang, L., Xu, L., Xia, L., Zhang, M., Zhang, M., Tang, M., Li, M., Wang, M., Li, M., Tian, N., Huang, P., Zhang, P., Wang, Q., Chen, Q., Du, Q., Ge, R., Zhang, R., Pan, R., Wang, R., Chen, R.~J., Jin, R.~L., Chen, R., Lu, S., Zhou, S., Chen, S., Ye, S., Wang, S., Yu, S., Zhou, S., Pan, S., Li, S.~S., Zhou, S., Wu, S., Ye, S., Yun, T., Pei, T., Sun, T., Wang, T., Zeng, W.,
  Zhao, W., Liu, W., Liang, W., Gao, W., Yu, W., Zhang, W., Xiao, W.~L., An, W., Liu, X., Wang, X., Chen, X., Nie, X., Cheng, X., Liu, X., Xie, X., Liu, X., Yang, X., Li, X., Su, X., Lin, X., Li, X.~Q., Jin, X., Shen, X., Chen, X., Sun, X., Wang, X., Song, X., Zhou, X., Wang, X., Shan, X., Li, Y.~K., Wang, Y.~Q., Wei, Y.~X., Zhang, Y., Xu, Y., Li, Y., Zhao, Y., Sun, Y., Wang, Y., Yu, Y., Zhang, Y., Shi, Y., Xiong, Y., He, Y., Piao, Y., Wang, Y., Tan, Y., Ma, Y., Liu, Y., Guo, Y., Ou, Y., Wang, Y., Gong, Y., Zou, Y., He, Y., Xiong, Y., Luo, Y., You, Y., Liu, Y., Zhou, Y., Zhu, Y.~X., Xu, Y., Huang, Y., Li, Y., Zheng, Y., Zhu, Y., Ma, Y., Tang, Y., Zha, Y., Yan, Y., Ren, Z.~Z., Ren, Z., Sha, Z., Fu, Z., Xu, Z., Xie, Z., Zhang, Z., Hao, Z., Ma, Z., Yan, Z., Wu, Z., Gu, Z., Zhu, Z., Liu, Z., Li, Z., Xie, Z., Song, Z., Pan, Z., Huang, Z., Xu, Z., Zhang, Z., and Zhang, Z.
\newblock Deepseek-r1: Incentivizing reasoning capability in llms via reinforcement learning, 2025.
\newblock URL \url{https://arxiv.org/abs/2501.12948}.

\bibitem[Eldan \& Li(2023)Eldan and Li]{tinystories}
Eldan, R. and Li, Y.
\newblock Tinystories: How small can language models be and still speak coherent english?, 2023.
\newblock URL \url{https://arxiv.org/abs/2305.07759}.

\bibitem[Fons et~al.(2024{\natexlab{a}})Fons, Kaur, Palande, Zeng, Balch, Veloso, and Vyetrenko]{fons2024evaluating}
Fons, E., Kaur, R., Palande, S., Zeng, Z., Balch, T., Veloso, M., and Vyetrenko, S.
\newblock Evaluating large language models on time series feature understanding: A comprehensive taxonomy and benchmark.
\newblock \emph{arXiv preprint arXiv:2404.16563}, 2024{\natexlab{a}}.

\bibitem[Fons et~al.(2024{\natexlab{b}})Fons, Kaur, Palande, Zeng, Balch, Veloso, and Vyetrenko]{fons2024evaluatinglargelanguagemodels}
Fons, E., Kaur, R., Palande, S., Zeng, Z., Balch, T., Veloso, M., and Vyetrenko, S.
\newblock Evaluating large language models on time series feature understanding: A comprehensive taxonomy and benchmark, 2024{\natexlab{b}}.
\newblock URL \url{https://arxiv.org/abs/2404.16563}.

\bibitem[Gao et~al.(2024)Gao, Koker, Queen, Hartvigsen, Tsiligkaridis, and Zitnik]{units}
Gao, S., Koker, T., Queen, O., Hartvigsen, T., Tsiligkaridis, T., and Zitnik, M.
\newblock Units: A unified multi-task time series model, 2024.
\newblock URL \url{https://arxiv.org/abs/2403.00131}.

\bibitem[Goswami et~al.(2024)Goswami, Szafer, Choudhry, Cai, Li, and Dubrawski]{moment}
Goswami, M., Szafer, K., Choudhry, A., Cai, Y., Li, S., and Dubrawski, A.
\newblock Moment: A family of open time-series foundation models, 2024.
\newblock URL \url{https://arxiv.org/abs/2402.03885}.

\bibitem[Gruver et~al.(2024)Gruver, Finzi, Qiu, and Wilson]{gruver2024largelanguagemodelszeroshot}
Gruver, N., Finzi, M., Qiu, S., and Wilson, A.~G.
\newblock Large language models are zero-shot time series forecasters, 2024.
\newblock URL \url{https://arxiv.org/abs/2310.07820}.

\bibitem[Gunasekar et~al.(2023)Gunasekar, Zhang, Aneja, Mendes, Giorno, Gopi, Javaheripi, Kauffmann, de~Rosa, Saarikivi, Salim, Shah, Behl, Wang, Bubeck, Eldan, Kalai, Lee, and Li]{textbooksneed}
Gunasekar, S., Zhang, Y., Aneja, J., Mendes, C. C.~T., Giorno, A.~D., Gopi, S., Javaheripi, M., Kauffmann, P., de~Rosa, G., Saarikivi, O., Salim, A., Shah, S., Behl, H.~S., Wang, X., Bubeck, S., Eldan, R., Kalai, A.~T., Lee, Y.~T., and Li, Y.
\newblock Textbooks are all you need, 2023.
\newblock URL \url{https://arxiv.org/abs/2306.11644}.

\bibitem[Hinton et~al.(2015)Hinton, Vinyals, and Dean]{hinton2015distillingknowledgeneuralnetwork}
Hinton, G., Vinyals, O., and Dean, J.
\newblock Distilling the knowledge in a neural network, 2015.
\newblock URL \url{https://arxiv.org/abs/1503.02531}.

\bibitem[Jin et~al.(2024)Jin, Zhang, Chen, Zhang, Liang, Yang, Wang, Pan, and Wen]{jin2024positionlargelanguagemodels}
Jin, M., Zhang, Y., Chen, W., Zhang, K., Liang, Y., Yang, B., Wang, J., Pan, S., and Wen, Q.
\newblock Position: What can large language models tell us about time series analysis, 2024.
\newblock URL \url{https://arxiv.org/abs/2402.02713}.

\bibitem[Oksendal(1998)]{biological}
Oksendal, B.
\newblock Stochastic differential equations, , an introduction with applications.
\newblock 1998.

\bibitem[Qwen et~al.(2025)Qwen, :, Yang, Yang, Zhang, Hui, Zheng, Yu, Li, Liu, Huang, Wei, Lin, Yang, Tu, Zhang, Yang, Yang, Zhou, Lin, Dang, Lu, Bao, Yang, Yu, Li, Xue, Zhang, Zhu, Men, Lin, Li, Tang, Xia, Ren, Ren, Fan, Su, Zhang, Wan, Liu, Cui, Zhang, and Qiu]{qwen2025qwen25technicalreport}
Qwen, :, Yang, A., Yang, B., Zhang, B., Hui, B., Zheng, B., Yu, B., Li, C., Liu, D., Huang, F., Wei, H., Lin, H., Yang, J., Tu, J., Zhang, J., Yang, J., Yang, J., Zhou, J., Lin, J., Dang, K., Lu, K., Bao, K., Yang, K., Yu, L., Li, M., Xue, M., Zhang, P., Zhu, Q., Men, R., Lin, R., Li, T., Tang, T., Xia, T., Ren, X., Ren, X., Fan, Y., Su, Y., Zhang, Y., Wan, Y., Liu, Y., Cui, Z., Zhang, Z., and Qiu, Z.
\newblock Qwen2.5 technical report, 2025.
\newblock URL \url{https://arxiv.org/abs/2412.15115}.

\bibitem[Rasul et~al.(2024)Rasul, Ashok, Williams, Ghonia, Bhagwatkar, Khorasani, Bayazi, Adamopoulos, Riachi, Hassen, Biloš, Garg, Schneider, Chapados, Drouin, Zantedeschi, Nevmyvaka, and Rish]{lagllama}
Rasul, K., Ashok, A., Williams, A.~R., Ghonia, H., Bhagwatkar, R., Khorasani, A., Bayazi, M. J.~D., Adamopoulos, G., Riachi, R., Hassen, N., Biloš, M., Garg, S., Schneider, A., Chapados, N., Drouin, A., Zantedeschi, V., Nevmyvaka, Y., and Rish, I.
\newblock Lag-llama: Towards foundation models for probabilistic time series forecasting, 2024.
\newblock URL \url{https://arxiv.org/abs/2310.08278}.

\bibitem[Spathis \& Kawsar(2023)Spathis and Kawsar]{spathis2023stephardestpitfallsrepresenting}
Spathis, D. and Kawsar, F.
\newblock The first step is the hardest: Pitfalls of representing and tokenizing temporal data for large language models, 2023.
\newblock URL \url{https://arxiv.org/abs/2309.06236}.

\bibitem[Team(2025)]{sky_t1_2025}
Team, N.
\newblock Sky-t1: Train your own o1 preview model within \$450.
\newblock https://novasky-ai.github.io/posts/sky-t1, 2025.
\newblock Accessed: 2025-01-09.

\bibitem[Wah et~al.(2017)Wah, Wright, and Wellman]{wah_wellman}
Wah, E., Wright, M., and Wellman, M.~P.
\newblock Welfare effects of market making in continuous double auctions.
\newblock 2017.

\bibitem[Williams et~al.(2018)Williams, Nangia, and Bowman]{williams2018broadcoveragechallengecorpussentence}
Williams, A., Nangia, N., and Bowman, S.~R.
\newblock A broad-coverage challenge corpus for sentence understanding through inference, 2018.
\newblock URL \url{https://arxiv.org/abs/1704.05426}.

\bibitem[Woo et~al.(2024)Woo, Liu, Kumar, Xiong, Savarese, and Sahoo]{moirai}
Woo, G., Liu, C., Kumar, A., Xiong, C., Savarese, S., and Sahoo, D.
\newblock Unified training of universal time series forecasting transformers, 2024.
\newblock URL \url{https://arxiv.org/abs/2402.02592}.

\bibitem[Xu et~al.(2024)Xu, Li, Tao, Shen, Cheng, Li, Xu, Tao, and Zhou]{xu2024surveyknowledgedistillationlarge}
Xu, X., Li, M., Tao, C., Shen, T., Cheng, R., Li, J., Xu, C., Tao, D., and Zhou, T.
\newblock A survey on knowledge distillation of large language models, 2024.
\newblock URL \url{https://arxiv.org/abs/2402.13116}.

\bibitem[Yuan et~al.(2023)Yuan, Yuan, Tan, Wang, and Huang]{yuan2023largelanguagemodelsperform}
Yuan, Z., Yuan, H., Tan, C., Wang, W., and Huang, S.
\newblock How well do large language models perform in arithmetic tasks?, 2023.
\newblock URL \url{https://arxiv.org/abs/2304.02015}.

\bibitem[Zhou \& Yu(2025)Zhou and Yu]{zhou2025llmsunderstandtimeseries}
Zhou, Z. and Yu, R.
\newblock Can llms understand time series anomalies?, 2025.
\newblock URL \url{https://arxiv.org/abs/2410.05440}.

\end{thebibliography}
\bibliographystyle{icml2025}


\end{document}